\documentclass[journal]{IEEEtran}
\usepackage{lettrine}   
\usepackage{xcolor}
\usepackage{amsmath,amsfonts,amssymb,amsthm}
\usepackage{graphicx}
\usepackage{array}
\usepackage{booktabs}
\usepackage[caption=false,font=normalsize,labelfont=sf,textfont=sf]{subfig}
\usepackage{algorithm}
\usepackage{algpseudocode}
\usepackage[backend=biber,style=ieee]{biblatex}
\usepackage{textcomp}
\usepackage{url}
\usepackage{hyperref}
\usepackage[T1]{fontenc}
\usepackage{cleveref}
\usepackage{orcidlink}

\algrenewcommand\algorithmicrequire{\textbf{Input:}}
\algrenewcommand\algorithmicensure{\textbf{Output:}}
\crefname{figure}{fig}{figures}
\Crefname{figure}{Fig}{Figures}
\usepackage{multirow}

\begin{document}

\title{Minkowski-MambaNet: A Point Cloud Framework with Selective State Space Models for Forest Biomass Quantification}

\author{Jinxiang Tu, Dayong Ren$^{*}$, Fei Shi, Zhenhong Jia, Yahong Ren, Jiwei Qin, Fang He
\thanks{Manuscript received September 12, 2025; This work was supported by the Key Research and Development Program of Xinjiang Uygur Autonomous Region under the project ``Research and Development of Key Technologies for Calculation, Measurement, and Digitized Management of Carbon Emission and Carbon Sink Indicators in the Energy Sector'' (No. 2022B01010), the National Natural Science Foundation of China (No. 62261053), the Tianshan Talent Training Project - Xinjiang Science and Technology Innovation Team Program (2023TSYCTD0012), and the Tianshan Innovation Team Program of Xinjiang Uygur Autonomous Region of China (2023D14012).}

\thanks{Jinxiang Tu,Fei Shi, Zhenhong Jia, and Jiwei Qin are with the School of Computer Science and Technology, the Key Laboratory of Signal Detection and Processing, and the Xinjiang Multimodal Intelligent Processing and Information Security Engineering Technology Research Center, Xinjiang University, Urumqi, 830046, China.}
\thanks{Dayong Ren (corresponding author e-mail:rdyedu@gmail.com) is with the National Key Laboratory for Novel Software Technology, Nanjing University, Nanjing 210023, China.}
\thanks{Yahong Ren is with Xinjiang Guanghui New Energy Co., Ltd, Urumqi, 830046, China.}
\thanks{Fang He is with Hongyousoft Co., Ltd, Karamay, 834000, China.}
}

\maketitle
\begin{abstract}
Accurate and robust quantification of forest biomass is crucial for effective forest management and global carbon cycle monitoring. Remote sensing, particularly airborne LiDAR, offers an unparalleled capability to capture detailed three-dimensional (3D) forest structures over large areas. However, directly estimating forest woody volume and aboveground biomass (AGB) from raw LiDAR point clouds remains a significant challenge due to the difficulty in capturing long-range dependencies, which are essential for distinguishing individual tree segments with similar local features. To address this, we introduce Minkowski-MambaNet, a novel deep learning framework designed for the direct estimation of forest woody volume and AGB from raw LiDAR point clouds. Our key innovation is the seamless integration of the Mamba principle, specifically its selective state space model (SSM), into a Minkowski-style point cloud network. This integration enables the model to effectively encode global context and long-range dependencies across the point cloud, thereby improving the differentiation of trees and segments. To enhance feature representation and accelerate convergence, we also incorporate skip connections, allowing the network to fuse multi-level features for robust regression and segmentation tasks. We evaluated Minkowski-MambaNet on airborne LiDAR data from the Danish National Forest Inventory. The results demonstrate that our proposed method significantly outperforms state-of-the-art methods based on point cloud statistics, yielding more accurate and robust estimations of woody volume and AGB. Importantly, our framework does not require a Digital Terrain Model (DTM) as a pre-processing step and exhibits robustness to artifacts at the boundaries of the evaluation area. This work presents a powerful tool for large-scale forest biomass dynamic analysis, providing a new pathway for LiDAR-based forest inventories.
\end{abstract}


\section{Introduction} \label{sec:introduction}
\IEEEPARstart{B}{iomass} estimation is a core task in ecology, climate change research, and sustainable forest management. It provides critical evidence for evaluating the carbon sequestration capacity of forests and plays a vital role in biodiversity monitoring, forest resource inventory, and ecological restoration planning. As forests represent the largest carbon pool in terrestrial ecosystems, the accurate quantification of their biomass has become a focal point of widespread international concern, especially in the context of global warming. According to the Intergovernmental Panel on Climate Change (IPCC), forest biomass accounts for approximately 40\% of global terrestrial carbon reserves, and its dynamic changes significantly impact the carbon cycle and greenhouse gas emissions\cite{c1,c2}. Over the past few decades, forest biomass estimation methods have gradually shifted from traditional field measurements to large-scale inversion supported by remote sensing technologies. However, challenges related to high costs, limited accuracy, and difficulties in large-scale application remain prevalent\cite{c4}.

Traditional biomass estimation methods primarily rely on field surveys and sample plot measurements, including destructive sampling and allometric equations \cite{c5}. While these methods offer high accuracy, they are time-consuming, labor-intensive, and difficult to apply to large-scale forested areas, particularly in remote regions with complex terrain. These methods are also constrained by the number of samples, failing to fully capture the spatiotemporal heterogeneity of forest structure and biomass, which can lead to estimation biases. In recent years, several studies have attempted to integrate satellite optical images with radar data for biomass inversion\cite{c51,c52}. Nevertheless, these 2D remote sensing data have obvious limitations in representing 3D forest structures, such as canopy occlusion and insufficient spatial resolution \cite{c7,c53}.

The advent of Light Detection and Ranging (LiDAR) technology has brought revolutionary progress to forest biomass estimation. Airborne LiDAR systems can acquire high-precision 3D point cloud data, directly capturing detailed information on vertical forest structures and finely depicting the spatial distribution of tree trunks, canopies, and branches \cite{c8,c54}. LiDAR point clouds contain rich geometric features, making them significantly superior to traditional 2D remote sensing data for estimating tree height, canopy volume, and biomass. However, the unstructured nature, uneven spatial density, and noise sensitivity of point clouds pose significant challenges. When processed by traditional Convolutional Neural Networks (CNNs), these characteristics result in high computational complexity and low efficiency. For large-scale forest scenarios, the massive volume and high-dimensional sparsity of point cloud data make it difficult for conventional algorithms to achieve efficient feature extraction and deep modeling.

In recent years, deep learning has achieved remarkable breakthroughs in point cloud processing, offering new insights for biomass estimation \cite{c14,c55}. Sparse Convolutional Neural Networks (e.g., MinkowskiNet) effectively reduce memory and computational overhead through sparse tensor computation, demonstrating excellent performance in tasks such as 3D semantic segmentation \cite{c15}. However, these methods rely on local convolution operations, which are inherently limited in capturing long-range dependencies within point clouds—a capability crucial for accurately estimating the biomass of trees with complex topological structures. The inability to model global context effectively often leads to a misinterpretation of similar local features, especially for distinguishing individual trees with distinct overall forms.

\begin{figure*}[!t]
    \centering
    
    \includegraphics[width=0.8\linewidth]{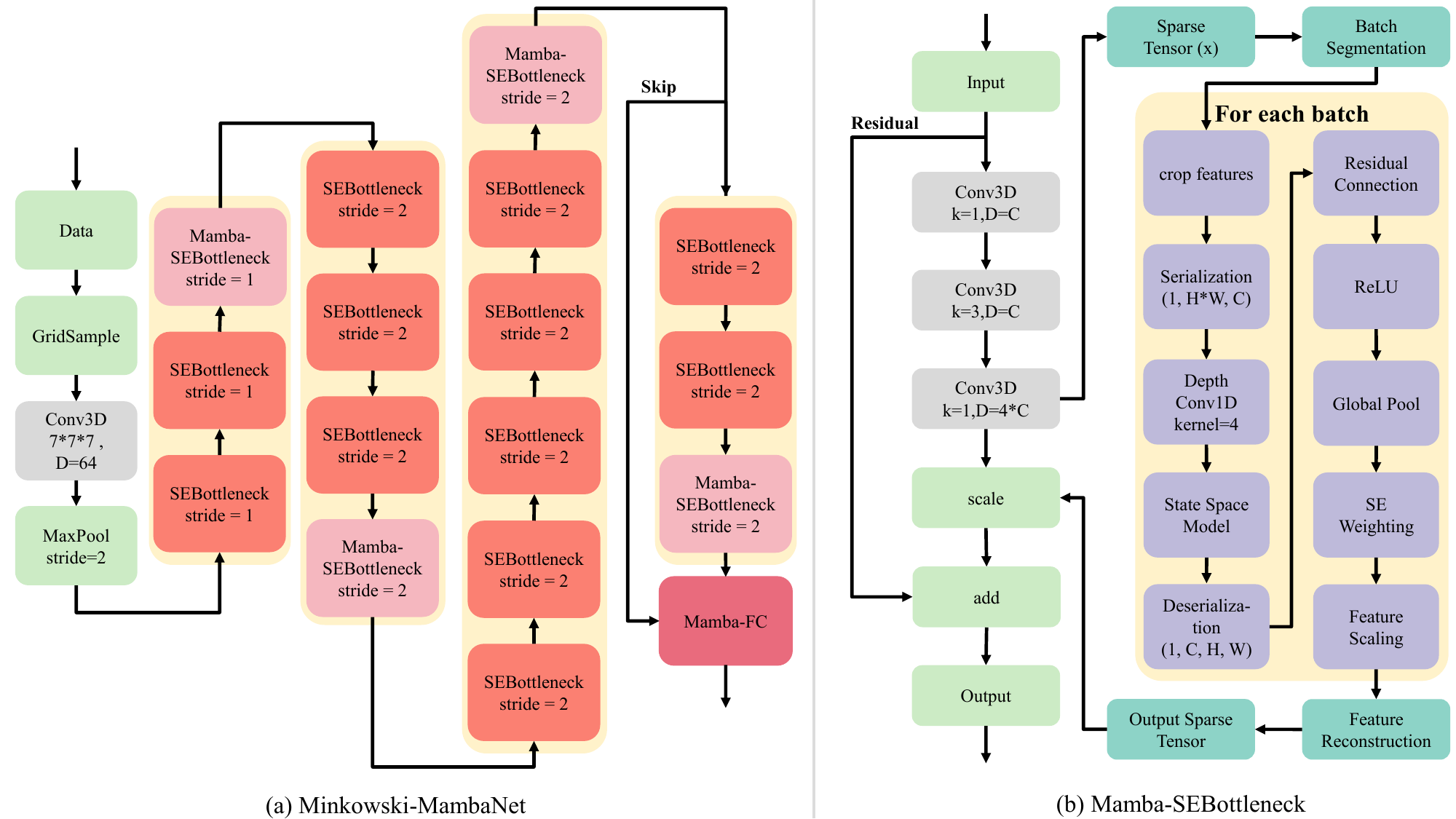}
    \vspace{0.5em} 
    
    \vspace{0.5cm} 
    
    \includegraphics[width=0.8\linewidth]{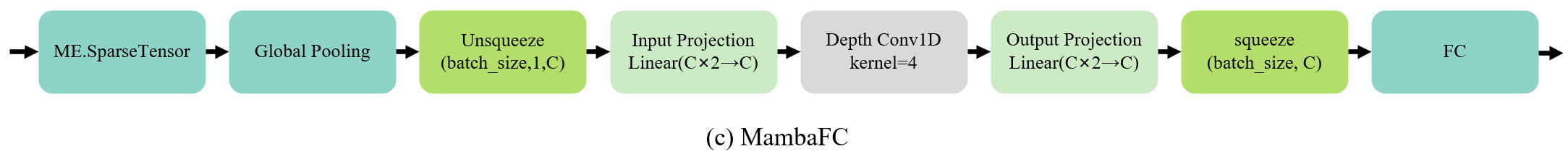}
    \vspace{0.5em} 
    
    \caption{In the MSENet50 architecture, there are a total of 16 Mamba-SEBlock, which are distributed across four stages with a layer count of (3, 4, 6, 3). Our strategy involves replacing the last SEBottleneck in each stage with a  Mamba-SEBottleneck, resulting in a configuration that includes 4  Mamba-SEBottleneck blocks and 12 SEBottleneck blocks.}
    \label{fig1}
\end{figure*}

To address this limitation and push the boundaries of direct biomass estimation from raw point clouds, this paper proposes a novel hybrid deep learning framework, Minkowski-MambaNet \ref{fig1}. The core innovation of this framework is the introduction of the  Mamba-SEBottleneck module, which combines the efficient sequence modeling capability of the Mamba Selective State Space Model (SSM) with Minkowski sparse convolution \cite{C19}. This integration breaks the bottleneck of traditional convolution in long-range dependency modeling. Mamba can process long sequences with linear complexity, a significant advantage over the quadratic computational complexity of self-attention mechanisms (e.g., Transformers) in high-dimensional point clouds. By integrating Mamba into the feature encoding process, the  Mamba-SEBottleneck can encode the hierarchical structure of trees (from trunks to branches and twigs) into continuous state sequences, enhancing the model's ability to distinguish between trees with similar local features but distinct global structures. This mechanism effectively addresses the challenge of decoupling local and global features, enabling the model to identify subtle structural differences critical for biomass estimation directly from plot-level point clouds. This approach obviates the need for cumbersome intermediate steps, such as explicit individual tree segmentation or species classification, which are prone to introducing errors.

To further mitigate the loss of shallow geometric details caused by progressive downsampling in sparse convolutional networks, this study also designs a Feature Fusion Modification Layer. Through a skip-connection mechanism, this layer fuses intermediate-layer features with deep-layer features. This structure not only accelerates convergence and alleviates gradient vanishing but also significantly enhances the model's ability to capture multi-scale features, thereby improving its generalization performance in complex forest stands.

This study validated the proposed method on an airborne LiDAR dataset from the Danish National Forest Inventory and compared it with state-of-the-art point cloud-based statistical methods. Experimental results demonstrate that Minkowski-MambaNet achieves higher accuracy in wood volume and AGB estimation. Additionally, our framework does not require preprocessing with a Digital Terrain Model (DTM) and exhibits strong robustness to noise in boundary regions. It is anticipated that this research will provide a powerful and effective tool for LiDAR-based dynamic monitoring of forest biomass.

\section{Related Work} \label{sec:relatedwork}

Light Detection and Ranging (LiDAR) point cloud data has become a fundamental technology for estimating Above-Ground Biomass (AGB) in forest ecosystems \cite{c23,c24}. It provides detailed 3D structural information of vegetation, essential for quantifying tree height, canopy volume, and trunk density. This section reviews the evolution, strengths, and limitations of both traditional and deep learning (DL) methods for point cloud-based biomass estimation.

\subsection{Traditional Methods for LiDAR-Based Biomass Estimation}

Traditional methods typically extract handcrafted geometric and statistical features from point clouds and relate them to field-measured biomass via regression. Common pre-processing steps include ground filtering, normalization, and vegetation segmentation. A widely used technique is the Canopy Height Model (CHM), which employs height metrics (e.g., mean, maximum, and percentile heights) to estimate biomass \cite{c26,c27}. Linear regression models applied to these metrics achieved reasonable accuracy in homogeneous forests \cite{c28}. Early foundational work by Nelson \cite{c49} and Nilsson \cite{c50} demonstrated the feasibility and accuracy of airborne LiDAR for forest parameter extraction.

Other approaches use point density analysis, linking the vertical distribution of returns to biomass \cite{c29}. At the individual tree level, Terrestrial Laser Scanning (TLS) produces high-density point clouds for detailed structural modeling. Methods such as voxel segmentation or cylinder fitting are used to extract tree parameters like Diameter at Breast Height (DBH) and volume, which are then converted to biomass using allometric models \cite{c31}. Recent improvements, such as the LiDAR Biomass Index (LBI), aggregate 3D structural metrics at plot and tree levels, improving performance in complex forests \cite{c32}. Portable LiDAR systems, including Handheld and UAV-based sensors, have extended these methods to finer scales \cite{c33,c34,c35}.

Despite their utility, traditional methods are limited by sensitivity to point density, reliance on manual feature design, poor generalization across diverse landscapes, and a heavy dependence on field data—making large-scale applications laborious and difficult to scale.

\subsection{Deep Learning for Point Cloud-Based Biomass Estimation}

Deep learning has transformed point cloud processing via end-to-end learning from raw 3D data, eliminating manual feature extraction. Specialized architectures such as PointNet \cite{C12}, PointCNN \cite{c36}, and voxel-based CNNs \cite{c37} excel at capturing hierarchical features and complex spatial relationships in vegetation.

Notable advances include deep regression frameworks that predict AGB or volume directly from point clouds. For example, Oehmcke et al. used convolutional layers on full point clouds to predict forest biomass, reducing RMSE by up to 17\%compared to traditional methods \cite{c38}. In agriculture, models like BioNet employ 3D CNNs for biomass regression and show scalability from plots to UAV imagery \cite{c39,c40}. Multimodal approaches integrating point clouds with optical imagery have also improved accuracy \cite{c41,c42}. Studies comparing deep and traditional methods, such as Random Forest, indicate superior performance of DL in temperate forests \cite{c44}.

A key remaining challenge is efficiently processing large-scale point clouds while capturing long-range dependencies. Local convolution-based methods struggle to model structures over large spatial extents, which is critical since tree biomass depends on global morphology. While Transformers effectively model long sequences with self-attention, their quadratic complexity hinders application to large point clouds \cite{c17}. Thus, developing novel architectures that efficiently process sparse 3D data and capture long-range context remains a central research challenge.

\section {Method} \label {sec:method}
\subsection{Sparse Convolutional Network Foundation} \label{subsec:minkowski_cnn}
The inherent sparsity of LiDAR point clouds has driven the development of deep learning frameworks specifically designed for sparse data, which offer significant memory and computational efficiency benefits \cite{c45}. Our work is built upon the Minkowski Engine \cite{c47}\cite{c15}, a powerful library for N-dimensional sparse convolution that has been widely adopted since its introduction following early work on Submanifold Sparse Convolutional Networks \cite{c46}.

Unlike traditional dense 3D convolutions that operate on a full voxel grid, the Minkowski Engine efficiently handles point clouds by voxelizing the input data into a sparse structure. Only voxels containing points are processed, which significantly reduces computational overhead. The core operation, Minkowski convolution, is formally defined as:
\begin{equation}
x_u^{\text{out}} = \sum_{i \in \mathcal{N}(u, \mathcal{C}^{\text{in}})} W_i x_{u+i}^{\text{in}}, \quad \text{for} \quad u \in \mathcal{C}^{\text{out}}.
\label{eq:minkowski_conv}
\end{equation}
Here, $\mathcal{C}^{\text{in}}$ and $\mathcal{C}^{\text{out}}$ are the sets of input and output sparse coordinates, respectively. $\mathcal{N}(u,\mathcal{C}^{\text{in}})$ represents the set of neighborhood offsets associated with position $u$ that are present in the input coordinates. $W_i$ are the learnable kernel weights, and $\mathbf{x}_{u+i}^{\text{in}}$ and $\mathbf{x}_{u}^{\text{out}}$ correspond to the input and output features.

In contrast, the formulation for conventional 3D convolution is presented as follows:
\begin{equation}
x_u^{\text{out}} = \sum_{i \in \mathcal{V}(K)} W_i x_{u+i}^{\text{in}}, \quad \text{for} \quad u \in \mathbb{Z}^3.
\label{eq:conventional_3d_conv}
\end{equation}
where $u \in \mathbb{Z}^{3}$ is a 3D coordinate, $K$ is the kernel size, and $\mathcal{V}(K)$ is the set of all offsets within the kernel. The key distinction is that Minkowski convolution only performs operations on non-empty coordinates, processing only the actually existing points and their neighborhoods. This selective operation drastically reduces computational complexity and memory requirements, making it highly suitable for efficiently handling large-scale forest point clouds containing millions of points.

\subsection{\textbf{Minkowski-MambaNet}}
\label{Minkowski-MambaNet}
This paper introduces Minkowski-MambaNet, a novel deep learning architecture designed to overcome two critical limitations of traditional convolutional neural networks (CNNs) in 3D point cloud-based forest biomass estimation: insufficient modeling of long-range dependencies and inadequate utilization of multi-scale features. The core innovation of Minkowski-MambaNet is the synergistic integration of efficient sparse 3D convolutional feature extraction with advanced state-space sequence learning mechanisms.

The network adopts MSENet50 \cite{c38} as its foundational backbone, which is then enhanced with two key modules: the  Mamba-SEBottleneck and the Feature Fusion Modification Layer. The  Mamba-SEBottleneck integrates the state-space model from Mamba, which significantly improves the network's capacity to model long-range dependencies and dynamic sequential patterns. Concurrently, the Feature Fusion Modification Layer establishes an efficient multi-scale feature transmission mechanism, ensuring the effective fusion and utilization of information across different abstraction levels. This hybrid architectural design enables Minkowski-MambaNet to more comprehensively perceive the complex structure of forest point clouds, leading to more accurate and robust biomass and volume regression.

\subsubsection{\textbf{ Mamba-SEBottleneck}}
\label{ Mamba-SEBottleneck}
Accurate biomass estimation from complex forest point cloud data critically depends on the effective modeling of long-range and dynamic spatial relationships. However, the Squeeze-and-Excitation (SE) modules in the MSENet50 architecture rely on global average pooling to generate static channel statistics. This approach struggles to capture the non-local, dynamically changing structural correlations inherent in forest point clouds, such as the hierarchical distribution from tree trunks to twigs or the spatial interactions between different canopy layers. The static nature and non-sequential processing of SE modules limit their ability to capture the fine-grained, dynamic dependencies necessary for a comprehensive understanding of complex spatiotemporal structures, thereby restricting the accuracy and generalization of biomass prediction.

To address these limitations, we introduce the  Mamba-SEBottleneck, which integrates the powerful sequence modeling capability of the Mamba state-space model into the sparse 3D convolution framework of the MinkowskiEngine. This design significantly enhances the network’s ability to capture long-range dependencies and dynamic sequence information while maintaining high computational efficiency.

The core of the  Mamba-SEBottleneck is the MambaSELayer, which replaces the standard SELayer in the SEBottleneck \cite{c38}. To adapt to the sparse and unordered nature of point cloud data (specifically, the MinkowskiEngine SparseTensor), we propose a heuristic transformation strategy: batch-wise, point features are flattened and then converted into an approximately square 2D grid through cropping and implicit padding. This process reconstructs the sparse point features into dense, pseudo-2D feature maps with local spatial correlations, making them compatible with Mamba's efficient sequence modeling.

The transformed pseudo-2D feature maps are converted into a sequential format and fed into the MambaBlock. The MambaBlock, which uses the Selective Scan mechanism, allows for dynamic information flow based on input content, efficiently capturing long-range dependencies and global context. This mechanism can implicitly encode the hierarchical structure of trees—from trunks to main branches and then to twigs—into continuous state sequences, effectively modeling the cumulative structural patterns that underpin biomass distribution. This approach overcomes the challenge of fusing local features with global context, a common limitation in traditional methods.

Features processed by the MambaBlock, which now incorporate long-range dependencies and dynamic sequence information, are passed through a sub-network of two fully connected layers, a GELU activation function, and a Sigmoid activation function to generate dynamic attention weights. These weights are then applied to the original sparse features (without cropping or reshaping), ensuring that the recalibration is based on a comprehensive understanding of both global and local point cloud contexts.

To better elucidate the advantages of the  Mamba-SEBottleneck, we compare its computational mechanism with that of the SEBottleneck \cite{c38}. The channel attention of the SEBottleneck is a static scaling operation on the input feature map $X \in \mathcal{R}^{C\times H\times W\times D}$, where $C$ is the number of channels and $H,W,D$ are spatial dimensions. This module first generates static channel statistics $z$ through global average pooling:
\begin{equation}
z = \frac{1}{\text{HWD}} \sum_{\mathrm{h}, \mathrm{w}, \mathrm{d}} \mathrm{X}_{\mathrm{h}, \mathrm{w}, \mathrm{d}}.
\end{equation}
This static nature limits its ability to model dynamic spatial relationships. The SEBottleneck then transforms these statistics via two fully connected layers to generate a weight vector $W$:
\begin{equation}
W = \sigma \left( W_{2} , \delta \left( W_{1} Z \right) \right).
\end{equation}
where $\delta$ is the ReLU activation function, $\sigma$ is the sigmoid activation function, and $W_{1},W_{2}$ are the weight matrices. The final output $\mathbf{X}^{\prime}$ is obtained by channel-wise scaling:
\begin{equation}
\mathbf{X}^{\prime} = \mathbf{w} \odot \mathbf{X}.
\end{equation}
The computational complexity of the SEBottleneck is primarily $O(C^2)$, arising from the matrix multiplications. Its static modeling mechanism makes it difficult to effectively capture non-uniform, long-range dependencies in point cloud data using a single set of channel weights.

In contrast, the  Mamba-SEBottleneck transforms an input sequence $\mathbf{X}\in \mathbb{R}^{\mathbf{T} \times \mathbf{C}}$ (where $\mathbf{T}$ is the sequence length, representing the number of points) via a state-space model. Its discrete-time update process unfolds as:
\begin{equation}
h_{t} = \sum_{k=1}^{t} A^{t-k} B x_{k} + A^{t} h_{0}.
\end{equation}
The output is a linear combination:
\begin{equation}
y_{t} = C h_{t} + D x_{t}.
\end{equation}
The computational complexity of Mamba is $O(TC)$, which scales linearly with the sequence length $\mathbf{T}$. While $\mathbf{T}$ can be significantly larger than $\mathbf{C}$ (e.g., a sub-plot may have 11,684 points, while $C$ is typically in the hundreds), Mamba's optimized implementation ensures a low practical computational overhead. The MambaSELayer generates attention weights by reshaping sparse tensors into a sequence and applying state-space updates. Unlike the SEBottleneck's static weights, these weights incorporate dynamic spatial information from the sequence, making them more adaptive to the non-uniformity of point clouds.

By leveraging these mechanisms, the  Mamba-SEBottleneck achieves efficient global modeling of long-range dependencies, for example, by effectively linking features from a tree trunk base to the morphological characteristics of its high-level canopy. This is crucial for an accurate understanding of the overall tree structure and biomass distribution. The dynamic, content-aware feature recalibration allows the network to better capture complex, non-linear spatial relationships. This design significantly enhances the discriminative ability of the network for point cloud structures, enabling it to distinguish features at different scales, such as branching patterns and crown outlines. By integrating the Mamba state-space model into the sparse convolutional framework, this module combines powerful sequence modeling with maintained computational efficiency, providing a robust foundation for high-precision forest biomass and volume estimation.

\subsubsection{\textbf{Feature Fusion Modification Layer: Multi-Scale Information Enhancement}}
\label{sec:feature_fusion_layer}
In sparse convolutional networks like MinkowskiNet, progressive downsampling is a common strategy to extract high-level semantic features. However, this process often leads to the dilution or loss of fine-grained geometric details captured by shallow layers. For forest biomass estimation, these details are critical, as the process relies on multi-scale structural information, including vertical canopy stratification, fine branch density, and macroscopic crown morphology. The loss of such information directly impairs the model's ability to understand complex forest stand structures, leading to a decline in prediction accuracy, particularly in scenarios with sparse point clouds or high heterogeneity.

To mitigate this information loss and enhance multi-scale feature transmission, we designed a Feature Fusion Modification Layer. This layer is a structured Skip Connection mechanism that facilitates the integration of multi-scale features. It selectively extracts features from an intermediate network layer (e.g., the output of the third stage of the backbone). These features typically contain medium-scale structural information, such as canopy outlines and main branch morphology, which are highly relevant for subsequent global feature aggregation and biomass prediction.

To effectively fuse these intermediate-layer features with the deep features from the final stage, the extracted features first undergo dimension alignment via a MinkowskiConvolution (with a 1$\times$1 sparse kernel) to match the channel count of the deep features. Subsequently, the intermediate-layer features are normalized using a skip-norm function. Finally, these processed features are added directly to the output of the network's final stage through a residual connection ($x=x+skip_x$). This weighted summation enables the model to learn the optimal integration of features from different abstraction levels, effectively fusing medium-scale structural details with global aggregated features. Beyond direct feature fusion, this skip connection also significantly improves gradient flow, alleviating the gradient vanishing problem and enabling the network to learn more complex feature patterns.

The Feature Fusion Modification Layer effectively complements the long-range dependency modeling capability of the  Mamba-SEBottleneck. This design optimizes the feature fusion strategy in sparse point cloud networks: it significantly mitigates information loss from downsampling while maintaining computational efficiency and enhancing the multi-level capture of geometric details. This synergistic mechanism allows the model to simultaneously perceive local details and global context, integrating them into a holistic decision-making process. By efficiently fusing multi-scale features, the proposed method demonstrates superior robustness, generalization, and prediction accuracy in remote sensing scenarios characterized by high forest structure heterogeneity and sparse data.

\begin{figure}[t!]
\centering
\includegraphics[width=0.45\textwidth]{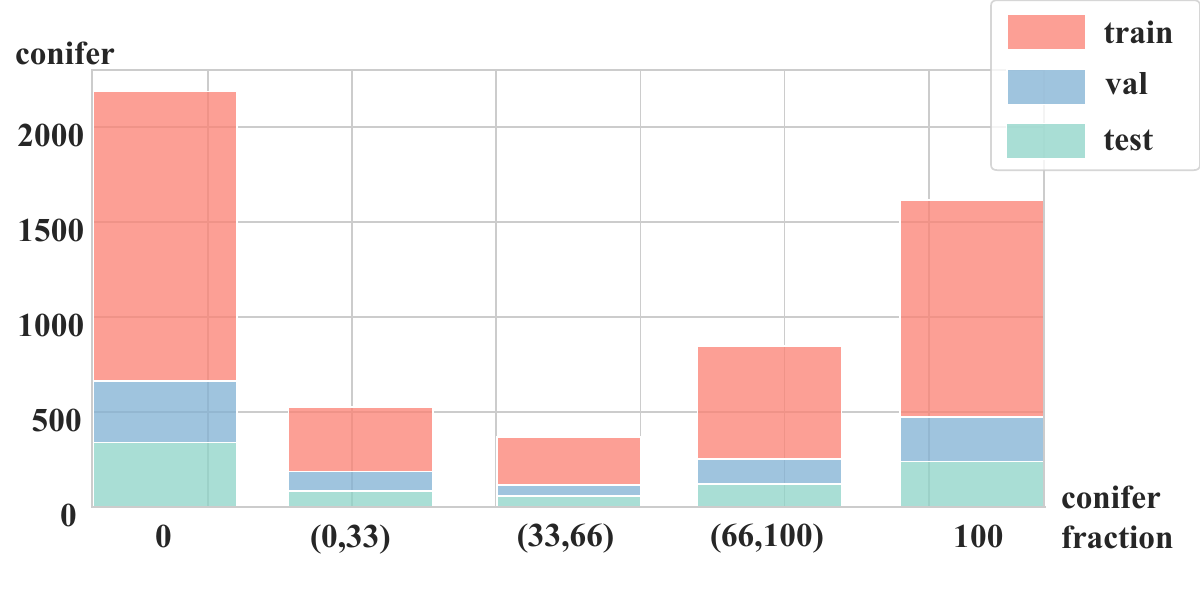}
\caption{Histogram of the proportion of coniferous forests in mixed forests across data splits.}
\label{fig2}
\end{figure}

\section{Experiment}
\subsection{Data Setup} \label{subsec:data_setup}
The dataset used in this study comprises two primary components: ground-truth forest inventory data and corresponding airborne LiDAR point cloud data. The ground-truth data, sourced from the Danish National Forest Inventory (NFI) from its 2013–2017 measurement cycle, provides measured values for standing wood volume (m$^3$$\cdot$ha$^{-1}$) and above-ground biomass (AGB) (Mg$\cdot$ha$^{-1}$). The NFI uses a systematic 2 km $\times$ 2 km grid system across the country, with one-fifth of the plots measured annually over a five-year cycle to ensure national representativeness. The dataset includes plots from coniferous-broadleaf mixed forests, with specific species compositions shown in Fig. \ref{fig2}.

The LiDAR point cloud data was obtained from the public 2014–2018 Danish National Airborne Laser Scanning Project. This nationwide campaign was primarily completed in 2014 and 2015, with an approximately one-fifth portion of the area rescanned in 2018. The data boasts a vertical and horizontal point accuracy of 5 cm and 15 cm (RMSE), respectively. We extracted point clouds corresponding to the NFI subplots, with each subplot containing an average of 11,684 points (standard deviation: 6,743) and an average point density of 16.7 points/m$^2$.

Both the NFI ground data and LiDAR point clouds underwent rigorous preprocessing consistent with established methods \cite{c38}. Key steps included:
\begin{itemize}
\item \textbf{Retention of Heterogeneity:} To enhance model robustness in real-world scenarios, we intentionally retained non-relevant elements such as trees outside the NFI subplots and non-forest man-made objects. This approach encourages the model to implicitly learn to identify and filter out such noise.
\item \textbf{Exclusion of Inconsistent Samples:} We systematically removed samples with significant temporal mismatches between LiDAR acquisition and ground measurements or with uncorrectable outliers to maintain data integrity and prevent spurious correlations.
\item \textbf{Minimum Height Threshold:} All samples without LiDAR points exceeding 1.3 meters in height were excluded. This threshold corresponds to the minimum height for Diameter at Breast Height (DBH) measurement, ensuring that all included samples contain relevant structural information for biomass estimation.
\end{itemize}

Following preprocessing, the final dataset consists of 6099 independent point cloud samples. To optimize the temporal alignment between LiDAR and ground measurements, the validation and test sets were strictly limited to samples where the LiDAR scan was conducted within one year of the biomass measurement. The training set allowed for a longer interval of up to nine years to increase sample diversity. The dataset composition and temporal constraints are detailed in Table \ref{R1}.

\begin{table}[t!]
\centering
\caption{Dataset composition for AGB and volume prediction}
\label{R1}
\begin{tabular}{lccc}
\hline
Dataset & Samples & Unique Subplots & Time Constraint \\
\hline
Training (total) & 4270 & 3938 & $\leq$ 9 years \\
\quad Time $\leq$ 1 year & 2636 & & \\
\quad Time $>$ 1 year & 1634 & & \\
Validation & 918 & 903 & $\leq$ 1 year \\
Test & 911 & 901 & $\leq$ 1 year \\
\hline
\end{tabular}
\vspace{1em}
\begin{minipage}{\linewidth}
\label{R1_notes}
\footnotesize
\textbf{Notes:}
\begin{itemize}[leftmargin=*, noitemsep]
\item The training set includes 2636 samples with a time interval of no more than one year and 1634 samples with a time interval longer than a year, located at 3938 unique subplot locations.
\item The validation set consists of 918 samples at 903 unique subplot locations.
\item The test set contains 911 samples at 901 unique subplot locations. Both validation and test sets have a maximum time distance of one year.
\item No site/plot used in the training or validation sets is included in the test set.
\end{itemize}
\end{minipage}
\end{table}

\subsection{Evaluation metrics}\label{Evaluation metrics}
Drawing on the work of Oehmcke et al. (2023) \cite{c38}, we evaluated models for predicting Aboveground Biomass (AGB) and timber volume using four key metrics: Root Mean Square Error (RMSE), Coefficient of Determination ($R^2$), Mean Absolute Percentage Error (MAPE), and Mean Bias (MB). Each metric offers distinct ecological and statistical insights into model performance.

\begin{equation}
\text{RMSE} = \sqrt{\frac{1}{N}\sum_{i=1}^{N}\left(y_i - f\left(x_i\right)\right)^2}.
\end{equation}

RMSE quantifies the square root of the mean of the squared discrepancies between predicted values $f(x_i)$ and observed values $y_i$, where $N$ is the sample size. It measures the dispersion of prediction errors, with lower values indicating higher model precision. In forest biomass estimation, RMSE is typically expressed in units of Mg$\cdot$ha$^{-1}$ (biomass) or m$^3$$\cdot$ha$^{-1}$ (timber volume), which directly assesses the model's utility in regional carbon sink accounting. However, its sensitivity to outliers necessitates complementing it with other metrics for a comprehensive evaluation.

\begin{equation}
R^2 = 1 - \frac{\sum_{i=1}^{N}\left(y_i - f\left(x_i\right)\right)^2}{\sum_{i=1}^{N}\left(y_i - \bar{y}\right)^2}, \quad \bar{y} = \frac{1}{N}\sum_{j=1}^{N} y_j.
\end{equation}

$R^2$ represents the proportion of data variability explained by the model, ranging from 0 to 1. Higher values denote a superior model fit. In forest research, an $R^2 > 0.8$ is often considered indicative of a high-precision model (Asner et al., 2011), suitable for large-scale carbon stock mapping. This metric is valuable for validating the explanatory power of predictive algorithms in heterogeneous forest environments.

\begin{table*}[t!]
    \centering
    \caption{Comparison of methods on the test set in terms of $R^{2}$, RMSE, MAPE, and MB. Note that $R^{2}$ and MAPE for biomass are identical to those for carbon storage. Zero biomass measurements were excluded from MAPE to avoid numerical issues. Units: Mg ha$^{-1}$ for AGB metrics, m$^3$ ha$^{-1}$ for timber volume.}
    \label{R2}
    \begin{tabular}{c l c c c c c c c c}
        \hline
        \textbf{Target} & \textbf{Model} & \multicolumn{2}{c}{\textbf{$R^{2}$}} & \multicolumn{2}{c}{\textbf{RMSE}} & \multicolumn{2}{c}{\textbf{MAPE(\%)}} & \multicolumn{2}{c}{\textbf{MB}} \\
        \hline
        & & med. & diff. & med. & diff. & med. & diff. & med. & diff. \\
        \hline
        \multirow{5}{*}[-2pt]{\textbf{AGB}}
        & Linear & 0.759 & -0.026 & 50.105 & -4.075 & 425.610 & -223.610 & 1.890 & -1.870 \\
        & RF & 0.742 & -0.043 & 51.000 & -4.970 & 625.521 & -423.521 & 1.470 & -1.457 \\
        & MSENet50 & 0.785 & 0.000 & 46.030 & 0.000 & 202.000 & 0.000 & 0.013 & 0.000 \\
        & PointNet & 0.766 & -0.019 & 50.683 & -4.653 & 386.459 & -184.459 & 1.011 & -0.998 \\
        & Minkowski-MambaNet & \textbf{0.810} & \textbf{+0.025} & \textbf{44.615} & \textbf{+1.416} & \textbf{163.150} & \textbf{+38.850} & \textbf{0.005} & \textbf{+0.008} \\
        \hline
        \multirow{5}{*}[-2pt]{\textbf{Volume}}
        & Linear & 0.766 & -0.008 & 87.350 & +4.052 & 185.640 & -47.525 & 4.710 & -3.546 \\
        & RF & 0.742 & -0.032 & 96.740 & -5.342 & 222.855 & -87.740 & 4.600 & -3.436 \\
        & MSENet50 & 0.774 & 0.000 & 91.398 & 0.000 & 138.115 & 0.000 & 1.164 & 0.000 \\
        & PointNet & 0.757 & -0.017 & 94.683 & -3.285 & 164.218 & -26.102 & 3.249 & -2.086 \\
        & Minkowski-MambaNet & \textbf{0.801} & \textbf{+0.027} & \textbf{85.860} & \textbf{+5.538} & \textbf{100.290} & \textbf{+37.825} & \textbf{0.119} & \textbf{+1.045} \\
        \hline
    \end{tabular}
    \vspace{0.5cm}
    \par\noindent\small\textit{Note: For $R^{2}$ (the higher the better), RMSE, MAPE, and MB (the lower the better), the difference is calculated as $\pm |(\text{Current model value}) - (\text{MSENet50 value})|$; the ``+'' sign indicates that the current model performs superiorly to MSENet50, while the ``-'' sign indicates that it performs inferiorly to MSENet50.}
\end{table*}

\begin{equation}
\text{MAPE} = \frac{100}{N} \sum_{i=1}^{N} \left| \frac{y_i - f(x_i)}{y_i} \right|.
\end{equation}

MAPE expresses prediction errors as percentages relative to observed values, making it a scale-independent metric. This is advantageous for comparing model performance across diverse forest types with varying biomass ranges. However, MAPE becomes unstable when $y_i$ approaches zero, necessitating careful application in sparse or young forest stands.

\begin{equation}
\text{MB} = \frac{1}{N} \sum_{i=1}^{N} \left( f(x_i) - y_i \right).
\end{equation}

MB measures systematic over- or underestimation. A positive MB indicates model overestimation, while a negative value reflects underestimation. In carbon sink assessments, minimizing MB is critical to avoid skewed regional carbon budget calculations. This metric is essential for diagnosing structural biases in remote sensing-based biomass models. These metrics collectively address distinct dimensions of model performance: RMSE quantifies overall precision, $R^2$ evaluates explanatory power, MAPE assesses relative error magnitude, and MB diagnoses directional bias.

\subsection{Experimental Results}
In this section, we present a detailed analysis of various models' performance for the forest biomass estimation task, based on the experimental data in Table \ref{R2}. The evaluation focuses on the prediction of Above-Ground Biomass (AGB) and wood volume, using a suite of key metrics: the coefficient of determination ($R^2$), Root Mean Square Error (RMSE), and Mean Absolute Percentage Error (MAPE). These metrics quantify the model's fitting ability, absolute prediction deviation, and relative error, respectively. We compared the performance of traditional statistical algorithms, including \textbf{Random Forest (RF)} and \textbf{Linear regression}, against established deep learning architectures such as \textbf{PointNet} and \textbf{MSENet50}, as well as our proposed \textbf{Minkowski-MambaNet}.

\subsubsection{Overall Model Performance Comparison}
The traditional models, RF and Linear regression, generally underperformed the deep learning approaches. While the Linear model's $R^2$ values (0.759 for biomass and 0.776 for volume) indicate a degree of linear correlation with target variables, they are surpassed by all deep learning models. Among the deep learning models, PointNet exhibited a statistical underperformance. Its Mean Absolute Percentage Error (MAPE) showed the largest discrepancy, primarily due to the amplified relative errors from samples with small target values. This limitation stems from its inability to fully leverage 3D topological information from unstructured point clouds, which can lead to overfitting.

\begin{figure}[t!]
\centering
\includegraphics[width=0.5\textwidth]{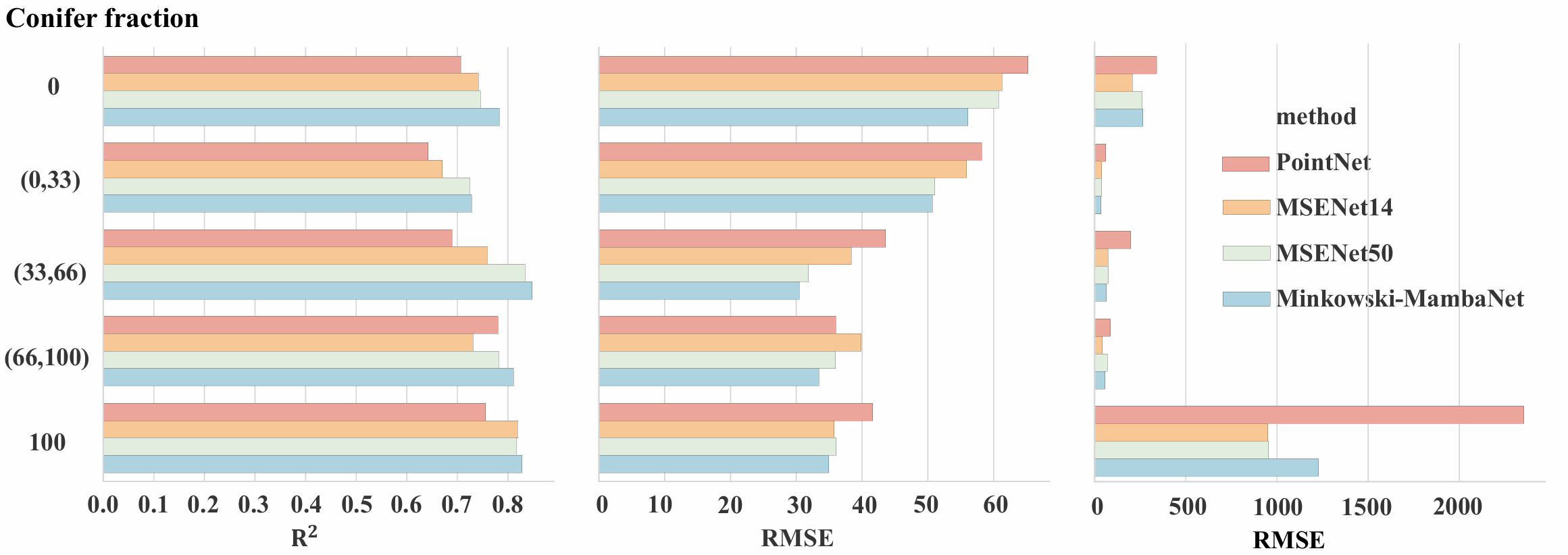} 
\caption{\(R^2\), RMSE, and MAPE of different components of biomass for coniferous and broadleaf trees on the test set. The columns represent\(R^2\)(the higher the better), RMSE (the lower the better), and MAPE (the lower the better). The corresponding values for wood volume are similar in nature.}
\label{fig3}
\end{figure}

\begin{figure*}[t!]

    \begin{minipage}{0.24\linewidth}
        \vspace{3pt}
        \centerline{\includegraphics[width=\textwidth]{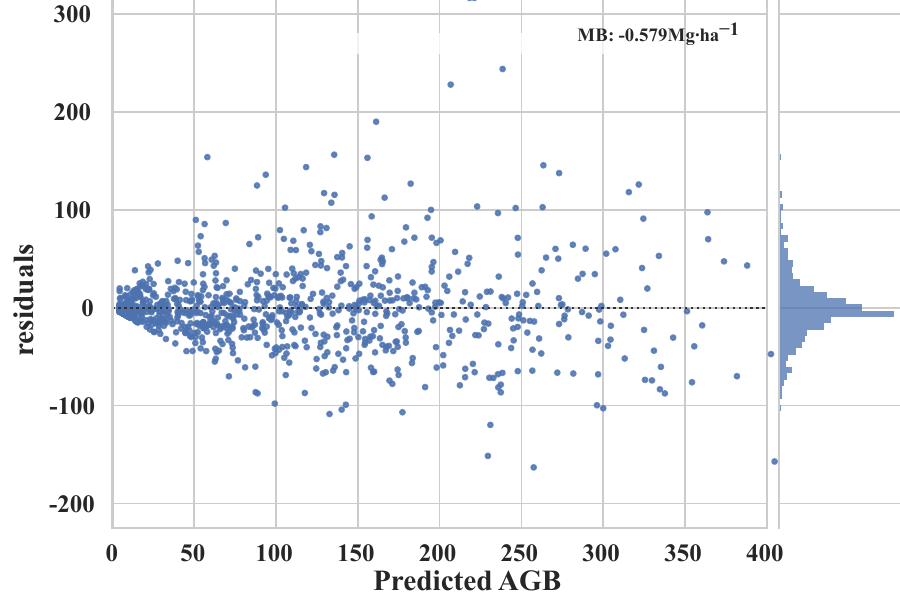}}
        \vspace{2pt}
        \centerline{\small\subfloat{PointNet}}
    \end{minipage}
    \hfill
    \begin{minipage}{0.24\linewidth}
        \vspace{3pt}
        \centerline{\includegraphics[width=\textwidth]{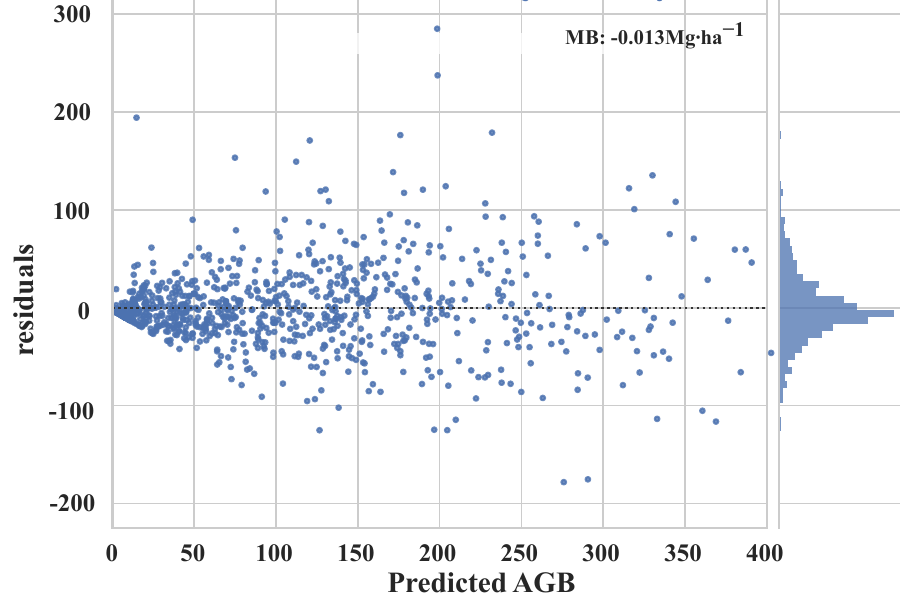}}
        \vspace{2pt}
        \centerline{\small\subfloat{MSENet50}}
    \end{minipage}
    \hfill
    \begin{minipage}{0.24\linewidth}
        \vspace{3pt}
        \centerline{\includegraphics[width=\textwidth]{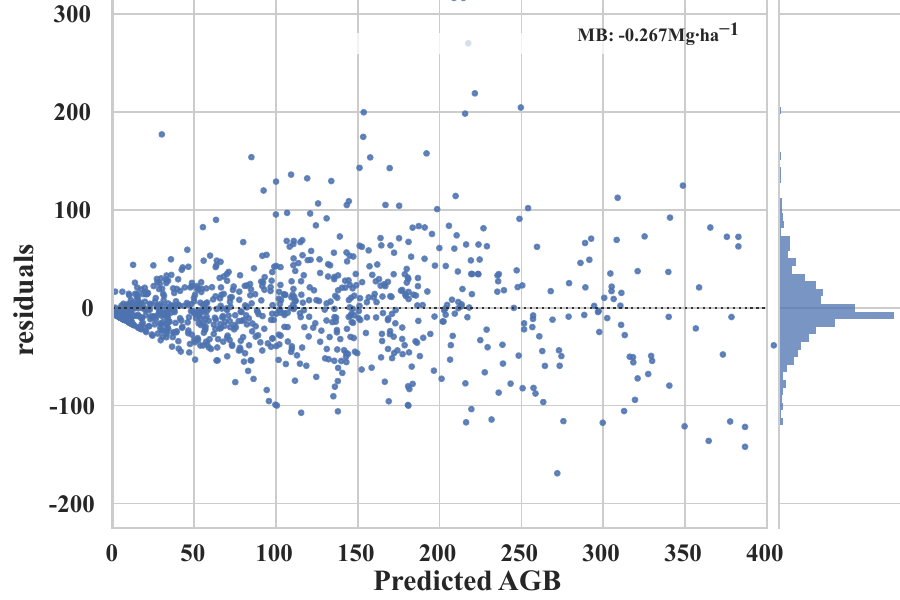}}
        \vspace{2pt}
        \centerline{\small\subfloat{MSENet14}}
    \end{minipage}
    \hfill
    \begin{minipage}{0.24\linewidth}
        \vspace{3pt}
        \centerline{\includegraphics[width=\textwidth]{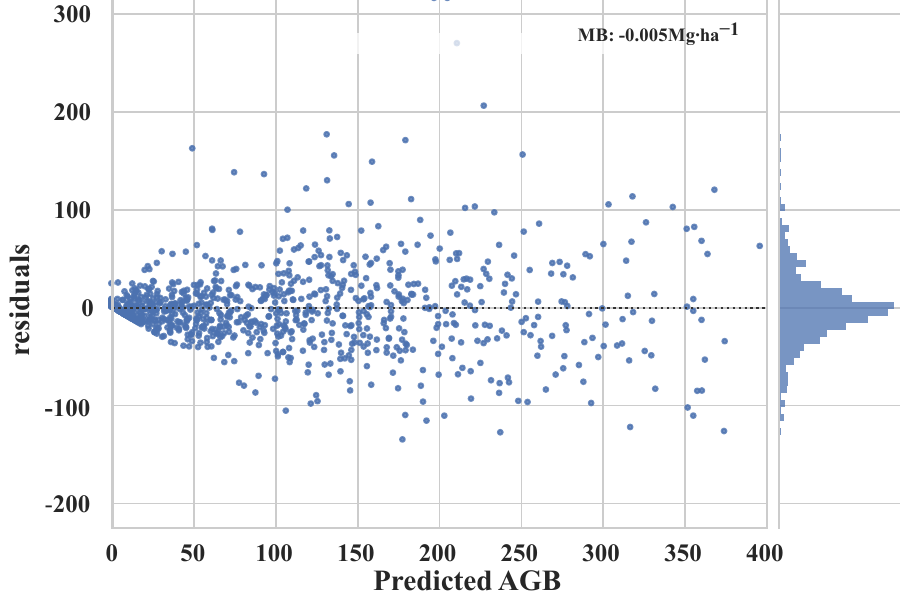}}
        \vspace{2pt}
        \centerline{\small\subfloat{MinkowskiMamba}}
    \end{minipage}  
    \caption{The test performance plots for the four methods (PointNet, MSENet50, MSENet14, and MinkowskiMamba) each consist of two parts: the biomass residual( left side) and the error distribution ( right side).}
    \label{fig5}
\end{figure*}

\begin{figure*}[t!]
	
	\begin{minipage}{0.245\linewidth}
		\vspace{3pt}
		\centering{\includegraphics[width=\textwidth]{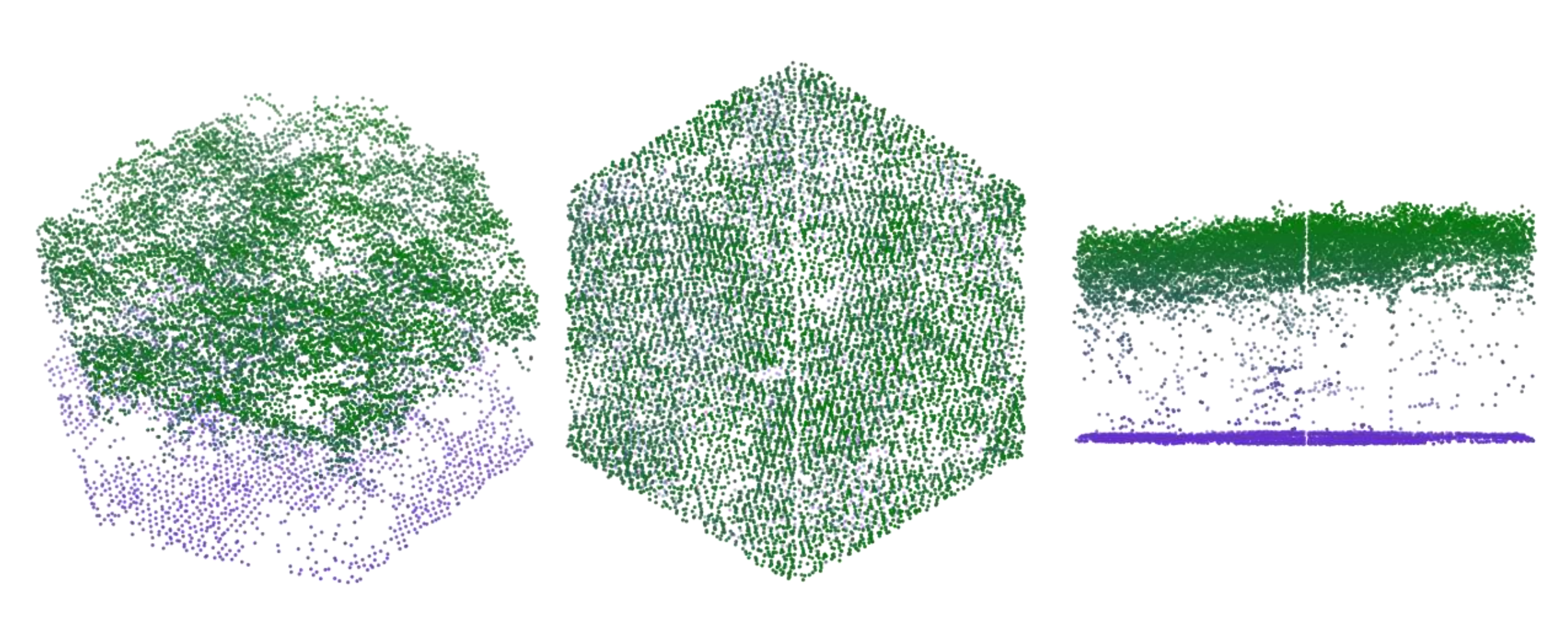}}
		\vspace{2pt}
		\centering{\small\subfloat{High}}
	\end{minipage}
	\begin{minipage}{0.245\linewidth}
		\vspace{3pt}
		\centering{\includegraphics[width=\textwidth]{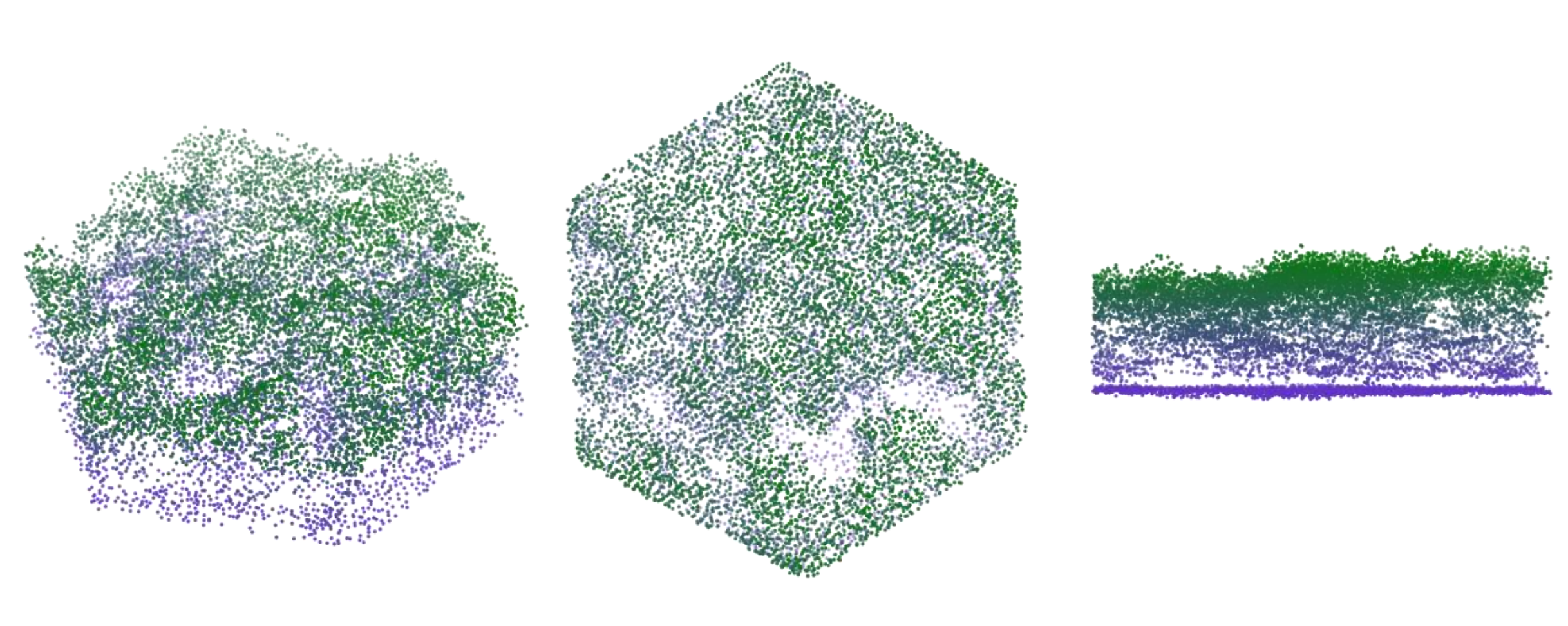}}
		\vspace{2pt}
		\centering{\small\subfloat{Short}}
	\end{minipage}
        \begin{minipage}{0.245\linewidth}
		\vspace{3pt}
		\centering{\includegraphics[width=\textwidth]{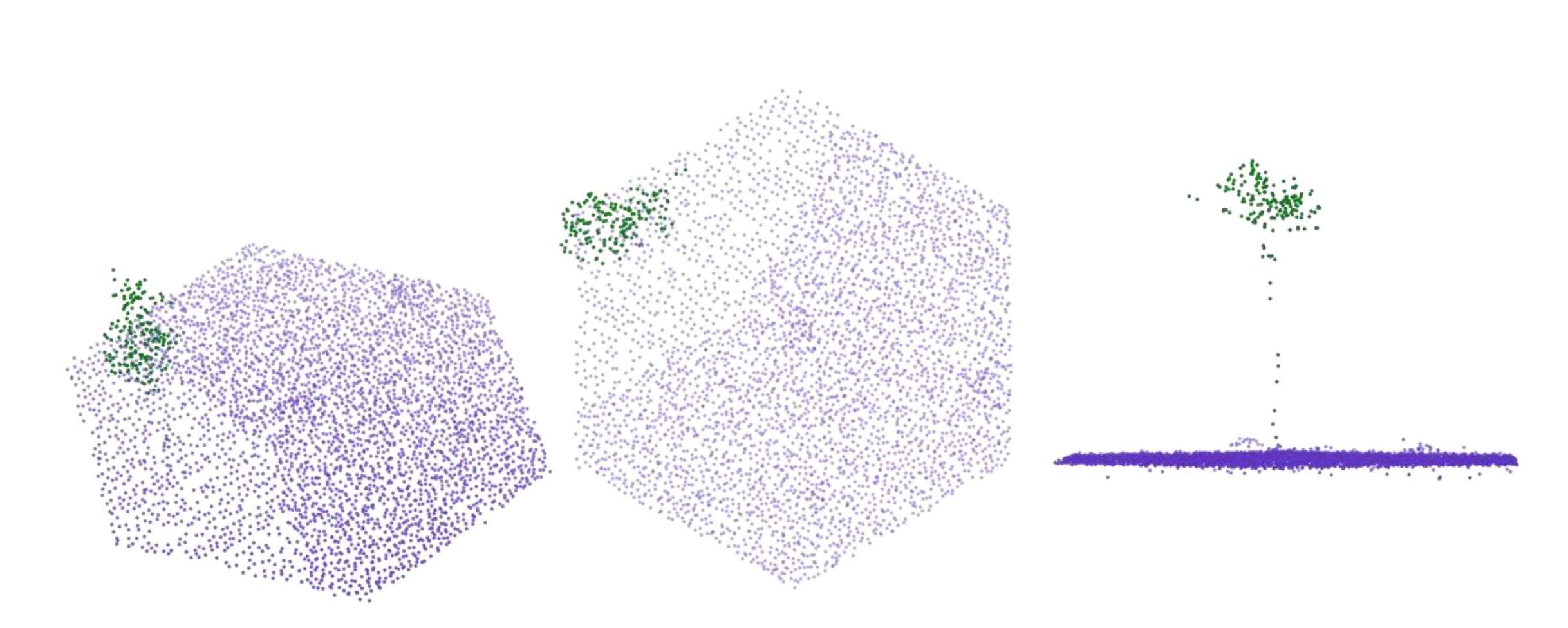}}
		\vspace{2pt}
		\centering{\small\subfloat{Single}}
	\end{minipage}
        \begin{minipage}{0.245\linewidth}
		\vspace{3pt}
		\centering{\includegraphics[width=\textwidth]{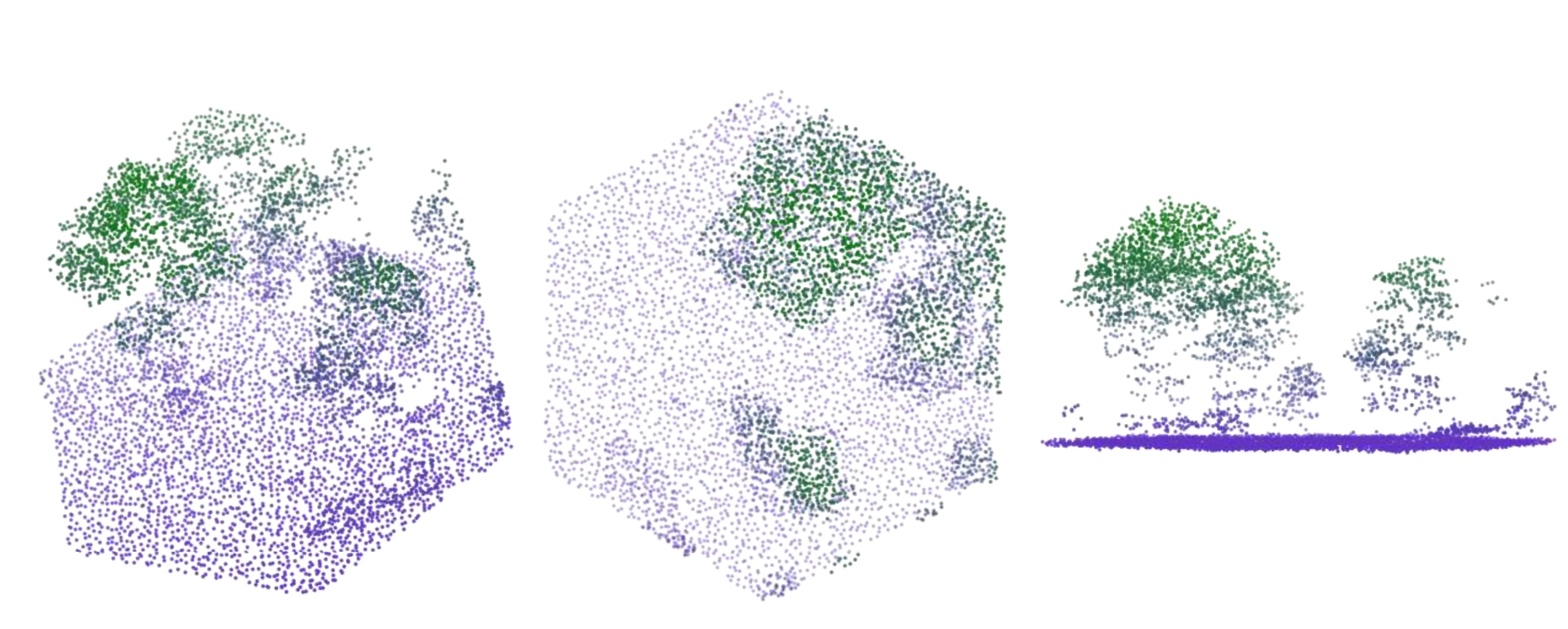}}
		\vspace{2pt}
		\centering{\small\subfloat{Small}}
	\end{minipage}
	\caption{More examples of AGB subplots with three perspectives in each group: isometric front view, top view, and side view. "High" refers to tall trees, and so on; "Short" refers to short trees; "Single" refers to a single tree; "Small" refers to small trees or sparse forests.}
    \label{fig6}
\end{figure*}

Both MSENet50 and Minkowski-MambaNet outperformed models based on precomputed features. MSENet50 achieved an $R^2$ of approximately 0.785, with an RMSE superior to PointNet but still inferior to our proposed framework. This suggests that while MSENet50's attention mechanism can enhance local feature selection, it struggles with the high computational overhead and limited generalization required for effective long-range dependency modeling.

Minkowski-MambaNet demonstrated the highest performance across all key metrics. It achieved an $R^2$ of \textbf{0.810} and an RMSE of \textbf{44.615 Mg$\cdot$ha$^{-1}$} for AGB, and an $R^2$ of \textbf{0.801} and an RMSE of \textbf{85.860 m$^3$$\cdot$ha$^{-1}$} for wood volume. This superior performance is attributed to its end-to-end learning mechanism, which seamlessly integrates 3D spatial features with global context. The framework's unique architecture, which fuses sparse convolution with custom Mamba blocks, enables efficient 3D feature extraction (e.g., processing canonical height sequences after global pooling), thereby significantly reducing prediction errors.

Furthermore, we analyzed model robustness by examining performance across varying proportions of coniferous and broadleaf forests using NFI data, as shown in Fig. \ref{fig3}. Our proposed Minkowski-MambaNet demonstrated excellent performance, with a notable advantage in mixed forest stands. It achieved the highest $R^2$ and lowest RMSE in stands with a coniferous proportion of 0.33–0.66, and it also performed exceptionally well in stands with a proportion of 0.66–1.00. This is because the model effectively addresses the challenge of decoupling local features and global context by encoding the complex, continuous hierarchical structure of forest point clouds into continuous states. This direct regression method bypasses error-prone intermediate steps like explicit individual tree segmentation or species classification, which significantly enhances the reliability of the predictions.

\begin{figure}[t!]
\centering
\begin{minipage}{0.49\linewidth}
	\vspace{3pt}
	\centerline{\includegraphics[width=\textwidth]{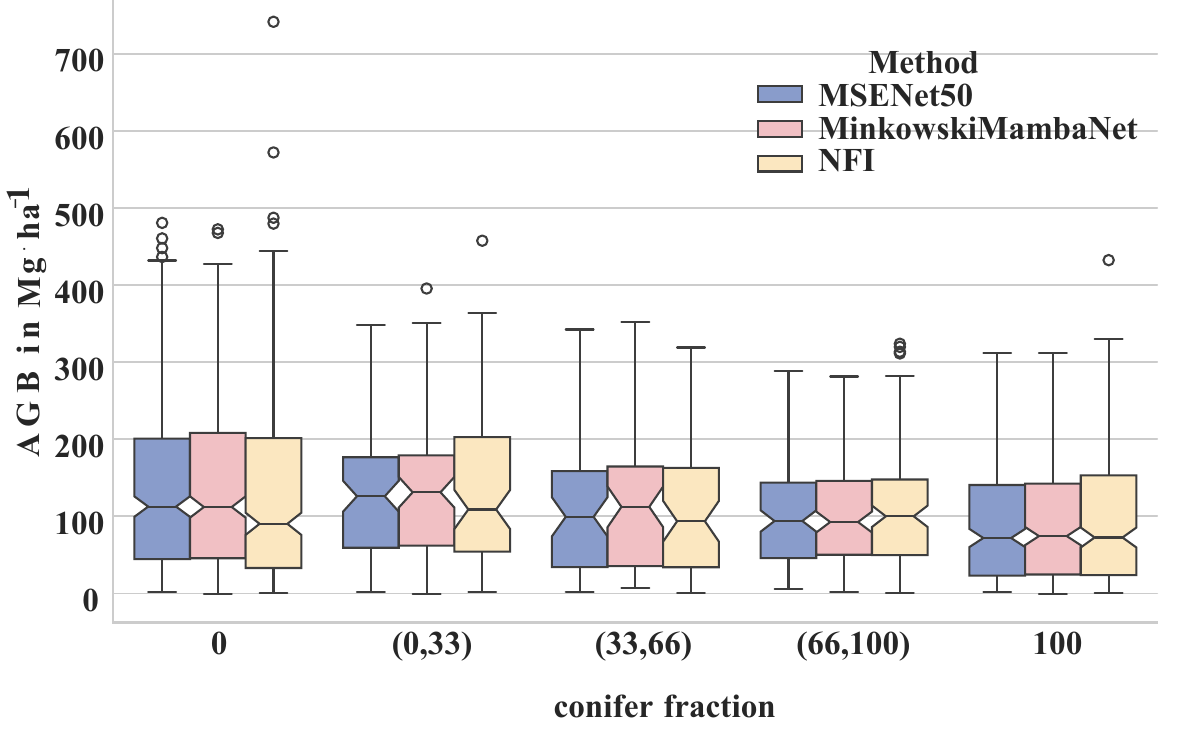}}
\end{minipage}
\begin{minipage}{0.49\linewidth}
	\vspace{3pt}
	\centerline{\includegraphics[width=\textwidth]{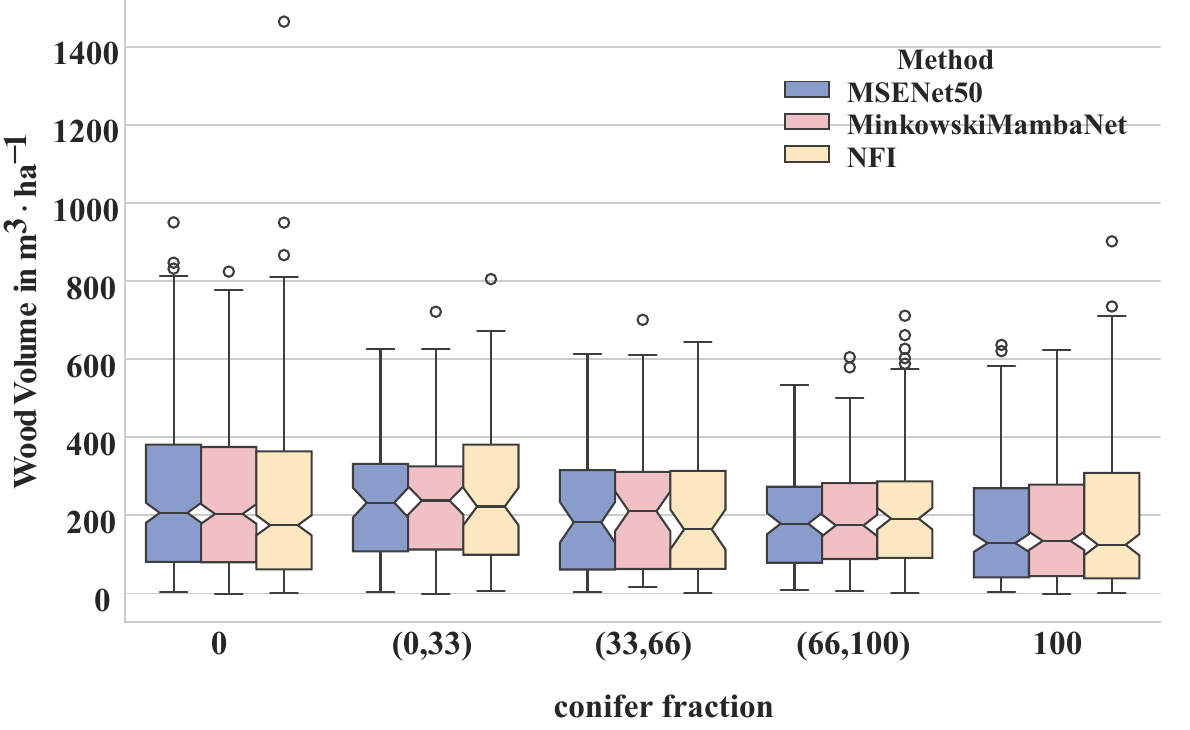}}	
\end{minipage}
\caption{Comparing value distribution of (a) AGB and (b) wood volume at different conifer fraction intervals with boxplots. The baseline (MSENet50) and MinkowskiMamba
model predictions are compared to the ground truth (NFI). Each box depicts the median in its center,  Outliers are shown as dots. The notches in the box represent the confidence interval around the median.}
\label{fig4}
\end{figure}

\subsubsection{Performance Visualization and Analysis}
As depicted in Fig. \ref{fig4}, the prediction distribution of Minkowski-MambaNet is noticeably more consistent with the actual ground-truth distribution compared to the MSENet50 baseline. The residual plots in Fig. \ref{fig5} further illustrate that for the Minkowski-MambaNet, MSENet50, PointNet, and MSENet14 models, the residuals tend to increase with the magnitude of the target variable. The symmetrical distribution of residuals around zero suggests that plot-level errors will partially offset during aggregation. However, a comparison of error distributions and average biases reveals that the error range of Minkowski-MambaNet is not yet as low as that of the best-performing Minkowski models.

Additional examples of subplots with particularly low or high AGB values are shown in Fig. \ref{fig6}. The model demonstrates excellent predictive capabilities for dense vegetation, as seen in the high-AGB examples of 420.9 and 372.1 Mg$\cdot$ha$^{-1}$, regardless of tree height. However, for sparse vegetation, the model exhibits certain limitations. For instance, the AGB for a single tree is slightly underestimated in the third example, while the prediction for sparse shrubs in the fourth example is slightly overestimated. This suggests that for sparse point clouds, the model finds it challenging to capture relevant features using common statistics, which contributes to the higher MAPE values observed.

\subsubsection{Ablation Experiments}
To verify the effectiveness of each module within the Minkowski-MambaNet model, we conducted ablation experiments to analyze the impact of the Minkowski- Mamba-SEBottleneck (MMB) and the Feature Fusion Modification Layer (FFM) on AGB and wood volume estimation. The experiments included the following configurations:
\begin{itemize}
    \item \textbf{MMB-Only}: This configuration tests the independent contribution of the Mamba state-space model by using only the Minkowski- Mamba-SEBottleneck.
    \item \textbf{FFM-Only}: This configuration assesses the effect of the feature fusion layer in enhancing feature transmission and alleviating gradient vanishing by using only the Feature Fusion Modification Layer.
    \item \textbf{Minkowski-MambaNet (Complete Model)}: This configuration represents the optimal architecture by integrating the MMB, FFM, and SENet channel attention.
\end{itemize}
The experiments were conducted on the Danish National Forest Inventory (NFI) dataset, with the workflow consistent with Section \ref{subsec:data_setup}. The evaluation metrics included $R^2$, RMSE, MAPE, and MB, as defined in Section \ref{Evaluation metrics}. Tables \ref{tab:ablation_agb} and \ref{tab:ablation_volume} present the ablation results for AGB and volume estimation, respectively.

Table \ref{tab:ablation_agb} shows that the complete Minkowski-MambaNet achieved the best AGB estimation performance, with an $R^2$ of \textbf{0.810}, an RMSE of \textbf{44.615 Mg$\cdot$ha$^{-1}$}, a MAPE of \textbf{163.15\%}, and an MB of \textbf{0.005 Mg$\cdot$ha$^{-1}$}. Compared to the other configurations, the full model exhibited a 0.37\%–1.25\% increase in $R^2$ and a 0.85\%–2.29\% decrease in RMSE, with an MB close to zero indicating minimal systematic bias.

\begin{table}[t!]
\centering
\caption{Ablation experimental results of AGB estimation}
\label{tab:ablation_agb}
\begin{tabular}{l c c c c}
\hline
\toprule
\textbf{Model} & \boldmath{$R^2$} & \textbf{RMSE } & \textbf{MAPE(\%)} & \textbf{MB } \\

\midrule
MMB       & 0.800        & 45.663    & 195.430    & 0.102     \\
FFM          & 0.807     & 45.001   & 215.648    & 0.106      \\
Minkowski-MambaNet & \textbf{0.810}  & \textbf{44.615}    & \textbf{163.150} & \textbf{0.005} \\

\bottomrule
\end{tabular}

\vspace{0.5em} 
\small
\textit{Note: This table mainly presents the ablation experiment results for AGB estimation, where MMB stands for Minkowski- Mamba-SEBottleneck and FFM stands for Feature Fusion Modification Layer. Bold values indicate the optimal results for each metric.}

\end{table}

Table \ref{tab:ablation_volume} presents the wood volume results, where the Minkowski-MambaNet again outperformed other configurations, achieving an $R^2$ of \textbf{0.800}, an RMSE of \textbf{85.860 m$^3$$\cdot$ha$^{-1}$}, a MAPE of \textbf{100.29\%}, and an MB of \textbf{0.119 m$^3$$\cdot$ha$^{-1}$}. Compared to the MMB-Only configuration ($R^2=0.791$, RMSE=87.545 m$^3$$\cdot$ha$^{-1}$), the complete model showed a 1.09\% increase in $R^2$ and a 1.92\% decrease in RMSE. These results confirm that the superior performance of Minkowski-MambaNet is due to the synergistic effect of its constituent modules.

\begin{table}[t!]
\centering
\caption{Ablation experimental results of wood volume estimation}
\label{tab:ablation_volume}
\begin{tabular}{l c c c c}
\hline
\toprule
\textbf{Model} & \boldmath{$R^2$} & \textbf{RMSE} & \textbf{MAPE (\%)} & \textbf{MB} \\

\midrule
MMB & 0.791 & 87.545 & 156.754 & 1.189 \\
FFM & 0.793 & 86.118 & 115.381 & 1.113 \\
Minkowski-MambaNet & \textbf{0.801} & \textbf{85.860} & \textbf{100.290} & \textbf{0.119} \\

\bottomrule
\end{tabular}

\vspace{0.5em} 
\small
\textit{Note: This table presents the ablation experiment results for the volume estimation task. MMB denotes Minkowski- Mamba-SEBottleneck, and FFM denotes Feature Fusion Modification Layer. Bold values indicate the optimal results for each metric.}
\end{table}

\section{conclusion}
In this study, we introduce Minkowski-MambaNet, a novel deep learning framework for the direct and robust quantification of forest woody volume and aboveground biomass (AGB) from raw airborne LiDAR point clouds. The framework effectively addresses two major challenges: capturing long-range dependencies and leveraging multi-scale features in complex forest ecosystems. Our core innovation lies in the  Mamba-SEBottleneck, which integrates the efficient Selective State Space Model (SSM) with Minkowski sparse convolutions. This allows the model to implicitly encode the hierarchical structures of trees into continuous states, a capability that is critical for distinguishing between trees with similar local features but distinct global forms. Additionally, a Feature Fusion Modification Layer with skip connections mitigates information loss during downsampling, ensuring the retention of fine-grained geometric details. Validated on the Danish National Forest Inventory dataset, Minkowski-MambaNet significantly outperforms state-of-the-art methods. It achieved a superior $R^2$ of 0.810 and a lower RMSE of 44.615 Mg·ha$^{-1}$ for AGB estimation. The framework operates directly on raw point clouds without a DTM, demonstrating strong robustness. Minkowski-MambaNet represents a significant advancement, offering a powerful and scalable tool for LiDAR-based forest inventories and dynamic monitoring.

\printbibliography

\newpage

\vfill

\end{document}